\documentclass[sigconf,screen=true]{acmart}
\renewcommand\footnotetextcopyrightpermission[1]{} 
\AtBeginDocument{%
  }

\usepackage{microtype}
\usepackage{graphicx}
\usepackage{subfigure}
\usepackage{booktabs} 
\usepackage{amsmath}

\usepackage{amssymb}
\usepackage{mathtools}
\usepackage{amsthm}
\usepackage{multirow}
\usepackage{xcolor}
\usepackage{colortbl}
\usepackage{bm}
\usepackage{hyperref}
\usepackage{enumitem}

\acmConference[KDD '25]{Proceedings of the 31st ACM
SIGKDD Conference on Knowledge Discovery and Data Mining V.2}{August 3--7,
2025}{Toronto, ON, Canada}

\begin{document}

\title[Offline Trajectory Optimization for Offline Reinforcement Learning]{Offline Trajectory Optimization for \\ Offline Reinforcement Learning}

\author{Ziqi Zhao}
\orcid{0009-0008-3011-5745}
\affiliation{%
  \institution{Shandong University}
  \city{Qingdao}
  \country{China}
}
\email{ziqizhao.work@gmail.com}

\author{Zhaochun Ren}
\orcid{0000-0002-9076-6565}
\affiliation{%
  \institution{Leiden University}
  \city{Leiden}
  \country{The Netherlands}
}
\email{z.ren@liacs.leidenuniv.nl}

\author{Liu Yang}
\orcid{0009-0007-7508-0964}
\affiliation{%
  \institution{Shandong University}
  \city{Qingdao}
  \country{China}
}
\email{yangliushirry@gmail.com}

\author{Yunsen Liang}
\orcid{0009-0003-5242-6716}
\affiliation{%
  \institution{Shandong University}
  \city{Qingdao}
  \country{China}
}
\email{yunsenliang@mail.sdu.edu.cn}

\author{Fajie Yuan}
\orcid{0000-0001-8452-9929}
\affiliation{%
  \institution{Westlake University}
  \city{Hangzhou}
  \country{China}
}
\email{yuanfajie@westlake.edu.cn}

\author{Pengjie Ren}
\orcid{0000-0003-2964-6422}
\affiliation{%
  \institution{Shandong University}
  \city{Qingdao}
  \country{China}
}
\email{renpengjie@sdu.edu.cn}

\author{Zhumin Chen}
\orcid{0000-0003-4592-4074}
\affiliation{
    \institution{Shandong University}
    \city{Qingdao}
    \country{China}
}
\email{chenzhumin@sdu.edu.cn}

\author{Jun Ma}
\orcid{0000-0003-0203-4610}
\affiliation{
    \institution{Shandong University}
    \city{Qingdao}
    \country{China}
}
\email{majun@sdu.edu.cn}

\author{Xin Xin}
\orcid{0000-0001-6116-9115}
\authornote{Corresponding author.}
\affiliation{%
  \institution{Shandong University}
  \city{Qingdao}
  \country{China}
}
\email{xinxin@sdu.edu.cn}
\renewcommand{\shortauthors}{Ziqi Zhao et al.}

\begin{abstract}
Offline reinforcement learning (RL) aims to learn policies without online explorations.
To enlarge the training data, model-based offline RL learns a dynamics model which is utilized as a virtual environment to generate simulation data and enhance policy learning.
However, existing data augmentation methods for offline RL suffer from (i) trivial improvement from short-horizon simulation; and (ii) the lack of evaluation and correction for generated data, leading to low-qualified augmentation.

In this paper, we propose \underline{\textbf{o}}ffline \underline{\textbf{t}}rajectory op\underline{\textbf{t}}imization for \underline{\textbf{o}}ffline reinforcement learning (OTTO).
The key motivation is to  conduct long-horizon simulation and then utilize model uncertainty to evaluate and correct the augmented data.  
Specifically, we propose an ensemble of Transformers, a.k.a. World Transformers, to predict environment state dynamics and the reward function. Three strategies are proposed to use World Transformers to generate long-horizon trajectory simulation by perturbing the actions in the offline data.
Then, an uncertainty-based World Evaluator is introduced to firstly evaluate the confidence of the generated trajectories and then perform the correction for low-confidence data.
Finally, we jointly use the original data with the corrected augmentation data to train an offline RL algorithm.
OTTO serves as a plug-in module and can be integrated with existing model-free offline RL methods. 
Experiments on various benchmarks show that OTTO can effectively improve the performance of representative offline RL algorithms, including in complex environments with sparse rewards like AntMaze.
Codes are available at \url{https://github.com/ZiqiZhao1/OTTO}.
\end{abstract}

\maketitle

\section{Introduction}
Offline reinforcement learning (RL) refers to training RL agents from pre-collected datasets without real-time interactions or online explorations on the true environment \cite{levine2020offline}.
This paradigm plays a crucial role in practical scenarios where collecting online interactions can be expensive or risky, such as healthcare~\cite{liu2020reinforcement,tang2022leveraging}, autonomous driving~\cite{yu2020bdd100k}, robotics~\cite{kalashnikov2018qtopt,rafailov2020offline,mandlekar2020iris,singh2020cog} and recommender systems~\cite{swaminathan2015batch,xiao2021general}. Standard RL methods often fail in such offline setting due to erroneous estimation of value functions~\cite{fujimoto2019off}.

\begin{figure}[t]
\begin{center}
\centerline{\includegraphics[width=0.9\columnwidth]{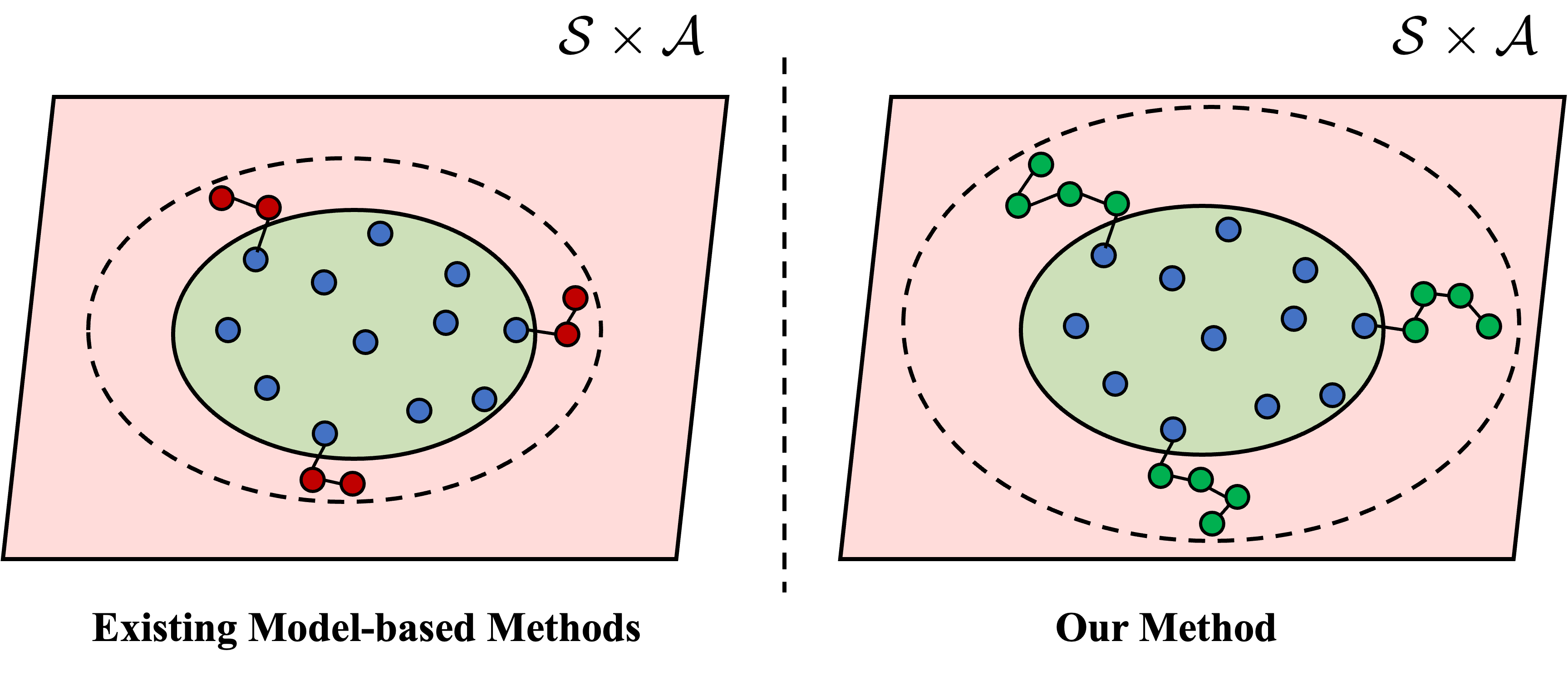}}
\caption{Comparison between existing model-based methods and OTTO. Blue dots refer to the original offline dataset. Red dots on the left represent the short-horizon rollouts from existing model-based methods, resulting in small generalization improvement near the original data. 
Green dots on the right represent the long-horizon trajectories generated by OTTO, which lead to a broader generalization range. }
\label{fig:long_short}
\end{center}
\vspace{-20pt}
\end{figure}

Existing offline RL methods can be classified into two categories: model-free methods and model-based methods.
Model-free offline methods incorporate conservatism into the value function estimation~\cite{fujimoto2019off,kumar2019stabilizing,fujimoto2018addressing,siegel2020keep,kumar2020conservative,kostrikov2021offlinefisher}.
For example, CQL~\cite{kumar2020conservative} adds a regularization term into the Q-function update. The conservatism downgrades the value function for unseen states and thus helps to avoid the over-estimation issue.  
As a result, the learned policy is constrained to near the offline data distribution. Besides, the overly conservatism could lead to underestimation of potentially good actions. Such issues lead to the poor generalization capability of model-free offline RL methods. 

\begin{figure*}[t]
    \centering
    \subfigure[]{
        \begin{minipage}[t]{0.3\linewidth}
            \centering
            \includegraphics[width=2in]{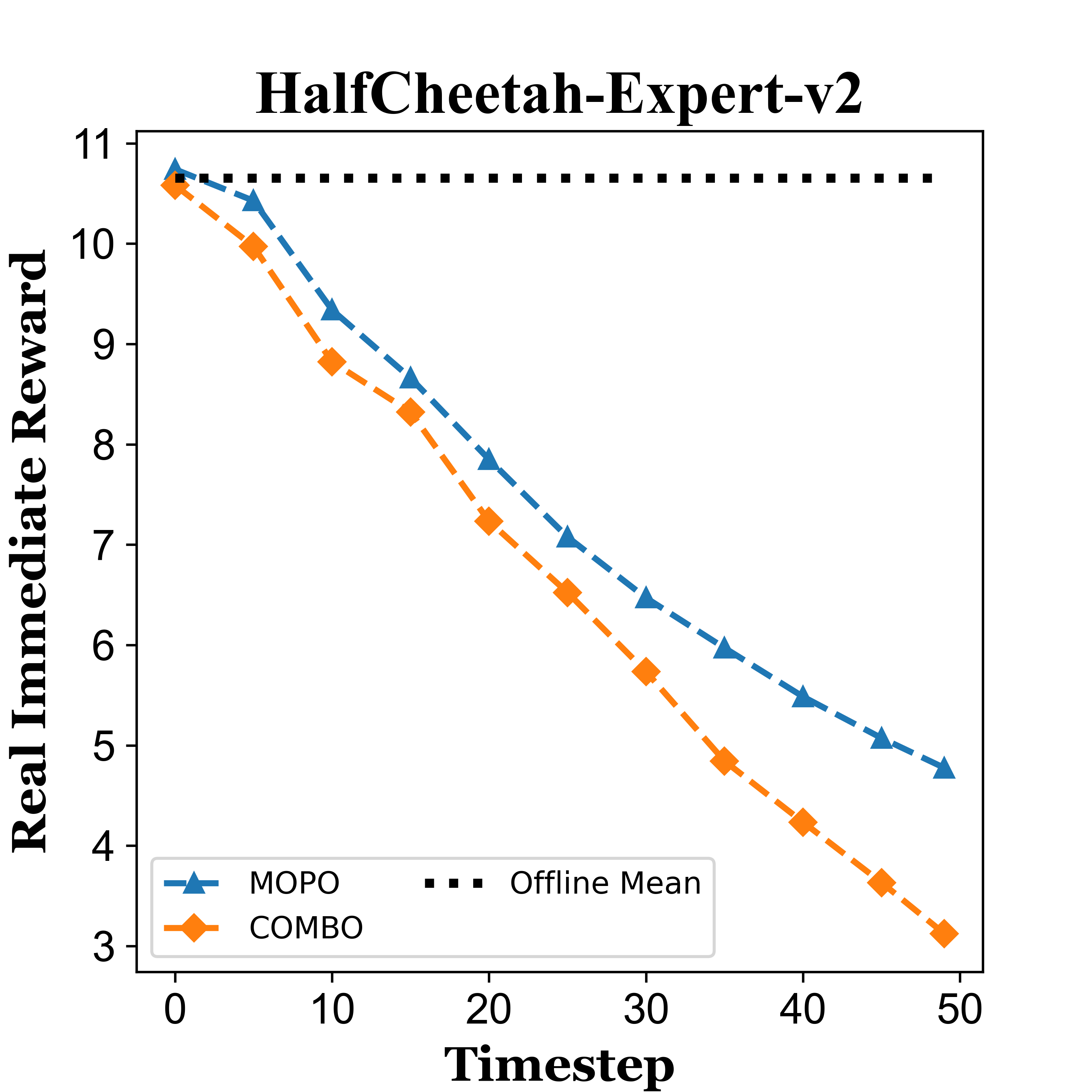}
        \end{minipage}
    }
    \subfigure[]{
        \begin{minipage}[t]{0.3\linewidth}
            \centering
            \includegraphics[width=2in]{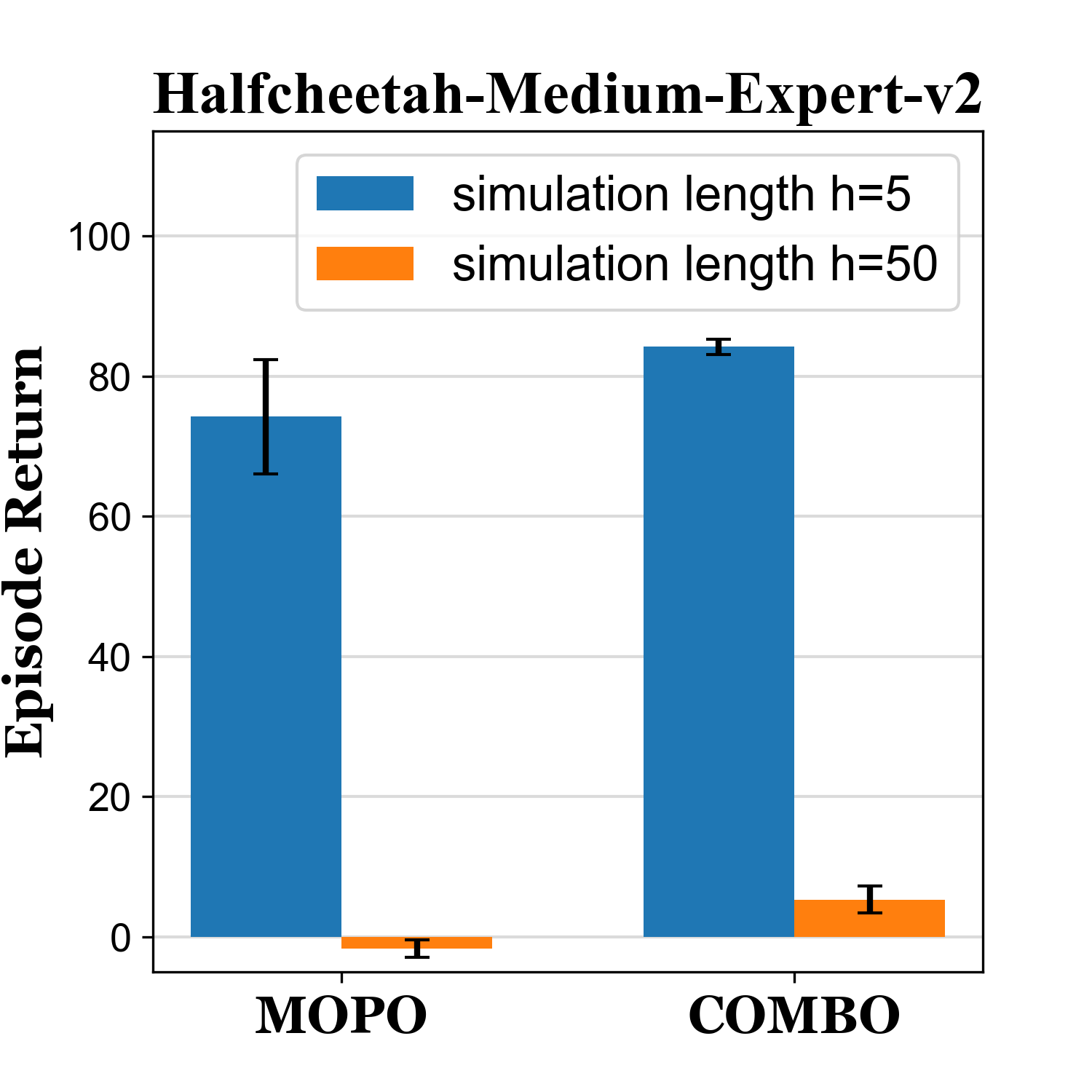}
        \end{minipage}
    }
    \subfigure[]{
        \begin{minipage}[t]{0.28\linewidth}
            \centering
            \includegraphics[width=2in]{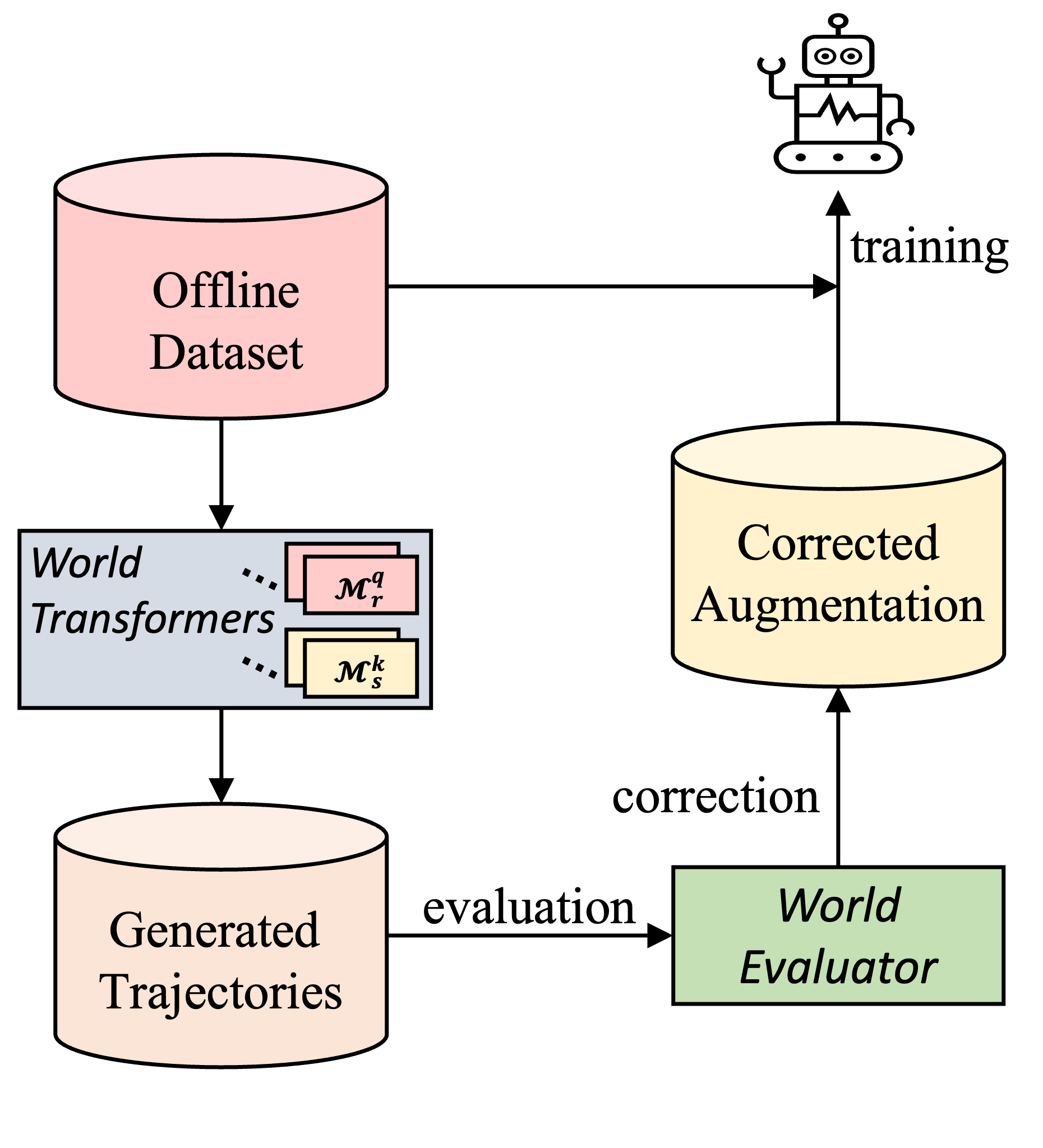}
        \end{minipage}
    }
    \vskip -0.1in
    \caption{
    (a) Average reward of each step. The average immediate reward of each interaction step in long simulation trajectories generated by two representative model-based methods MOPO \cite{yu2020mopo} and COMBO \cite{yu2021combo}. The reward becomes lower for longer steps.
    (b)  Policy performance of MOPO and COMBO with short and long simulations over 5 different seeds. Long simulation with lower reward downgrades the performance. 
    (c) {The framework of the proposed OTTO.}}
    \label{fig:intro}
\vspace{-8pt}
\end{figure*}


To enhance the generalization capability of offline RL, a viable solution is to generate additional training data. Model-based offline RL algorithms learn a dynamics model based on the offline dataset, which is then utilized as a simulated environment to generate new data  ~\cite{yu2020mopo,kidambi2020morel,yu2021combo}. 
However, we found that the augmented data generated by existing methods is of low qualification, resulting in trivial improvement for policy learning. 
Specifically, most existing model-based RL approaches only perform short-horizon model rollouts, resulting in marginal generalization improvement only near the offline data, as shown in Figure~\ref{fig:long_short}. 
Besides, when we perform the simulation of long-horizon trajectories using the environment model, the 
average reward of each interaction step often becomes lower for longer steps, as shown in Figure \ref{fig:intro}(a). As a result, the performance of the learned policy suffers a sharp decline, as shown in Figure \ref{fig:intro}(b).
Additionally, although there exist some works utilizing the powerful generative model to augment the offline data, these data augmentation methods lack the evaluation and correction for the generated trajectories. Consequently, the augmented trajectories of existing methods could be unreliable, leading to sub-optimal performance.
To summarize, existing data augmentation methods for offline RL suffer from (1) trivial improvement from short-horizon rollout simulation and (2) the lack of evaluation and correction for generated trajectories.

To address the aforementioned issues, we propose  \underline{\textbf{o}}ffline \underline{\textbf{t}}rajectory op\underline{\textbf{t}}imization  for \underline{\textbf{o}}ffline reinforcement learning (\textbf{OTTO}). 
Figure \ref{fig:intro}(c) demonstrates the framework of OTTO.
To address the first issue, we propose to learn the state dynamics transition and the reward function through an ensemble of Transformers \cite{vaswani2017attentionTFM}, a.k.a. World Transformers, based on the offline dataset. 
To reduce the computational cost, we employ a cyclic annealing schedule to reset the environment model's learning rate and preserve model snapshots, thereby obtaining an ensemble of environment models from a single training process.
Then, three strategies are proposed to use World Transformers to generate long trajectory simulation by perturbing the actions in offline data. 
The generalization and long sequence modeling capability of World Transformers, together with the prior knowledge in the offline dataset, ensure that the generated long-horizon trajectories are high-reward and thus help the model to learn potential good actions.
To address the second issue, we propose an uncertainty-based World Evaluator to 
conduct the evaluation and correction for generated trajectories. The main idea is that if World Transformers are not confident about the state transition and reward, a conservative punishment is introduced to correct the simulated trajectories to avoid potential overestimation.
Finally, original offline data together with the corrected augmentations are jointly used to train an offline RL algorithm. 
Note that OTTO serves as a plug-in module and can be integrated with a wide range of model-free offline RL methods to further enhance the learned policy. 
The main contributions are as follows:
\begin{itemize}
    \item We propose World Transformers, which utilize an ensemble of Transformers to learn state dynamics transition and reward functions with good generalization and long sequence modeling capability.
    \item We propose three trajectory simulation strategies, which generate high-qualified long-horizon trajectory simulations by using the World Transformers.
    \item We propose an uncertainty-based World Evaluator to evaluate and correct the generated trajectories, which further enhances the reliability of the augmented data.
    \item Experiments on various benchmarks show that OTTO can effectively improve the performance of representative offline RL algorithms, including in complex environments with sparse rewards like AntMaze.  
\end{itemize}

\section{Related Work}
Offline RL aims to learn policies from a pre-collected, static dataset of pre-collected trajectories. It is important to various domains in which online explorations are expensive to collect, e.g., healthcare~\cite{liu2020reinforcement,tang2021model}, autonomous driving~\cite{yu2020bdd100k,zhu2021survey}, NLP~\cite{jaques2019way,jaques2020human}, and recommender systems~\cite{swaminathan2015batch,chen2022off}.

\subsection{Model-free offline RL} 
Existing model-free offline RL methods are mainly based on restricting the target policy to be near the behavior policy \cite{fujimoto2019off,kostrikov2021offlineimplicit,fujimoto2021minimalist,wu2019behavior,siegel2020keep}, including  incorporating value pessimism \cite{kumar2020conservative,xie2021bellman,kostrikov2021offlinefisher}, importance sampling \cite{liu2019off_is,nachum2019algaedic_is,sutton2016emphatic_is}, and uncertainty punishment for value functions \cite{sinha2022s4rl_unc,an2021uncertainty_unc,kumar2019stabilizing}. 
Recently, \citet{chen2021decision,janner2021offline}
proposed to cast offline RL into a sequence modeling problem and use Transformers to solve it in a supervised manner.
However, these methods are limited to the support of offline data distribution.
Since the state distribution for practical usage might differ from that of the behavior policy, model-free offline RL encounters the distributional shift issue, leading to sub-optimal performance when evaluating in real environments. 
In this paper, we propose to use World Transformers and World Evaluator to augment high-qualified data and further enhance the performance of these existing methods.




\subsection{Data augmentation for offline RL}\label{subsec:da4offline}
Existing model-based offline RL methods \cite{yu2020mopo,kidambi2020morel,yu2021combo,rigter2022rambo, zhan2022modelbased, swazinna2021overcoming, rafailov2020offline, matsushima2020deploymentefficient,lowrey2018plan,schrittwieser2021online} learn a model of environment and generate synthetic rollouts to improve the policy. However, we found the augmented rollouts generated by these methods are extremely short-horizon (mostly with simulation length $h=1$), resulting in limited generalization improvement. 
Although BooT\cite{wang2022bootstrapped} uses a Transformer to model the environment and generate data for bootstrapping, it still can only generate short trajectories with length $h=1$ in most cases and fails to integrate effectively with other methods.
On the contrary, the proposed OTTO aims to generate long-horizon trajectories with high quality through World Transformers, providing more potential for better generalization. 

Besides, some works propose to utilize the powerful diffusion-based model to augment the offline data \cite{li2024diffstitch,lu2024synthetic}. For instance, SER \cite{lu2024synthetic} learns the environment transition tuples and samples new tuples through diffusion models to augment the data. DiffStitch \cite{li2024diffstitch} selects a low-reward trajectory and a high-reward trajectory, then generates a sub-trajectory to stitch them together by the diffusion model. However, these data augmentation methods lack the evaluation and correction for the generated trajectories.
Consequently, the augmented trajectories of existing methods could be unreliable. In this paper, the proposed OTTO introduces a World Evaluator to evaluate the uncertainty of generated trajectories and further uses a conservative punishment to correct low-confidence trajectories.

Some online RL works \cite{zhang2024storm,micheli2022transformers,chen2022transdreamer} also use Transformers to model the environment, but these works are primarily focused on utilizing Transformers to improve sampling efficiency, and they cannot be directly applied to offline RL. 
In contrast, our work leverages Transformers to improve the generalization capability of offline RL by generating long-horizon trajectories.
The problem we aim to solve is fundamentally different from that of online RL.


\section{Methodology}
This section presents the details of OTTO. 
We first describe the task in section \ref{subsec:task}. 
Then, the World Transformers are described in section \ref{subsec:worldmodel}.
The long-horizon trajectory generation is described in section \ref{subsec:trajectorygen}.
Finally, the World Evaluator is described in section \ref{subsec:worldevaluator}.

\subsection{Task description}
\label{subsec:task}
A Markov Decision Process (MDP) is defined by the tuple 
\begin{math}
<\mathcal{S},\mathcal{A},
\end{math}
\begin{math}
\mathit{T},r,\mu_0,\gamma>,
\end{math}
where $\mathcal{S},\mathcal{A}$ refer to state space and action space, respectively. 
A policy $\pi(a|s)$ defines a mapping from state $s\in\mathcal{S}$ to a probability distribution over action $a\in\mathcal{A}$.
Given a $(s,a)$ pair, $T(s'|s,a)$ and $r(s,a)$ represent the distribution of the next state $s'$ and the obtained immediate reward, respectively. $\mu_0$ is the initial state distribution and $\gamma \in (0,1)$ is the discount factor for future reward.
The goal of RL is to find an optimal policy $\pi^*={\text{argmax}}_{\pi}\mathop{\mathbb{E}}\limits_{\pi,\mathit{T}} [\sum_{t=0}^\infty \gamma^tr(s_t,a_t)|s_0\sim\mu_0]$ that maximizes the expected cumulative reward. $t$ denotes the interaction step.

\begin{figure*}[ht]
\begin{center}
\centerline{\includegraphics[width=0.95\linewidth]{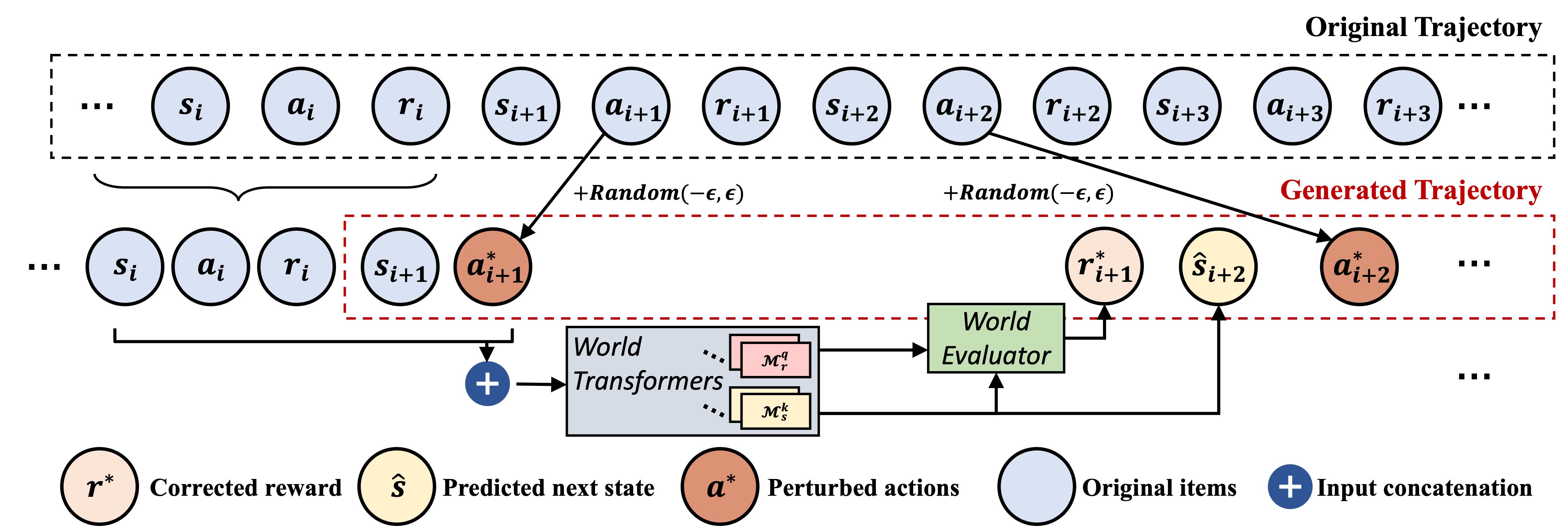}}
\caption{Trajectories augmentation of OTTO. Given an original trajectory and the augmentation step $t=i+1$,  random noise of range $(-\epsilon,\epsilon)$ is introduced to perturb $a_{i+1}$ and forms a new action $a_{i+1}^*$. Then the corresponding $r_{i+1}^*$ and $\hat{s}_{i+2}$ can be inferred from World Transformers and World Evaluator. This process is repeated for $h$ times to generate the new trajectory for augmentation.}
\label{fig:framework}
\end{center}
\vspace{-20pt}
\end{figure*}

Conventional RL performs online exploration through sampling $(s,a)$ according to the target policy $\pi$, which could be expensive under practical usage.  
To address this issue, offline RL aims to learn a policy from a static dataset $\mathcal{D}$, which contains trajectories $\tau=\{s_{t},a_{t},r_{t}\}_{t=0}^{|\tau|}$ pre-collected with an unknown behavior policy. However, offline RL algorithms often fail to tackle unseen states and actions during online inference in a real environment. 
In this work, we aim to improve the performance of offline RL through generating
high-qualified long-horizon simulated trajectories, which enlarge the observed states and actions during  model training.
$\mathcal{D}_{aug}$ is used to denote the augmented dataset.
$\delta=|\mathcal{D}_{aug}|/|\mathcal{D}|$ is used to denote the augmentation ratio. The  process for trajectory augmentation of OTTO is visualized in Figure \ref{fig:framework}.

\subsection{World transformers}\label{subsec:worldmodel}
To perform trajectory augmentation, 
an environment simulator needs to be implemented to model the state transition distribution and the reward function.  
We propose to use Transformers to build the simulator, a.k.a. World Transformers. Precisely, World Transformers consist of State Transformers and Reward Transformers to predict the next state and reward, respectively.
It has been shown that Transformers can effectively learn from long input sequences and produce promising outputs \cite{chen2021decision,devlin2018bertNLP,brown2020languageNLP,raffel2020exploringNLP}.


\textbf{Model inputs.} The key factors to construct the environment model inputs are: (1) the input should contain all the factors determining the environment dynamics; (2) the input should enable World Transformers to learn from continuous sequential interactions; (3) the input should be as simple as possible to avoid unnecessary noise. Considering that the environment provides feedback simply based on the agent's action and the current state, the model input for World Transformers contains multiple continuous state and action tuples $(s,a)$. Formally, we define the model input as
\begin{equation}\label{eql:tra}
inputs=\{\cdots,s_t,a_t,s_{t+1},a_{t+1},\cdots\}
\end{equation}
Both State Transformers and Reward Transformers take the above inputs for environment simulation.

\textbf{Environment modeling.} 
We use State Transformers and Reward Transformers to model the next state transition and the reward function, respectively. 
Specifically, State Transformers are defined as $\mathbf{M}_s={\{\mathcal{M}_s^k|k\in\{1,2,...K}\}\}$, and Reward Transformers are defined as $\mathbf{M}_r={\{\mathcal{M}_r^q|q\in\{1,2,...Q}\}\}$. $K$ and $Q$ are the number of models, respectively.  
$\mathcal{M}_s^k$ and $\mathcal{M}_r^q$
are the $k$-th State Transformer and the $q$-th Reward Transformer.
We introduce a learnable embedding matrix to represent the interaction step $t$, which is similar to the positional embedding in language modeling. Noticing that GPT-series models have achieved remarkable success across various NLP tasks \cite{radford2018improving,brown2020languageNLP,radford2019language}, we use the similar architecture to model the states transition and reward function.  At timestep $t$, $\mathcal{M}_s^k$ and $\mathcal{M}_r^q$ model the following distributions:
\begin{equation}\label{equ:sinput1}
\hat{s}_{t+1}^k\sim p_{\mathcal{M}_{s}^k}(\hat{s}_{t+1}|{\{{s},{a}\}}_{t-{L}_{s}+1}^t)
\end{equation}
\begin{equation}\label{equ:sinput2}
\hat{r}_{t}^q\sim p_{\mathcal{M}_{r}^q}(\hat{r}_{t}|{\{{s},{a}\}}_{t-{L}_{r}+1}^t),
\end{equation}
where ${\{{s},{a}\}}_{t-L+1}^t$ represents the model input sequence $\{{s}_{t-L+1},$
${a}_{t-L+1},{s}_{t-L+2},{a}_{t-L+2},...,{s}_{t},{a}_{t}\}$ as shown in Eq.(\ref{eql:tra}). 
$L_s,L_r$ represent the hyperparameters that define the length of the model inputs for $\mathcal{M}_{s}^k,\mathcal{M}_{r}^q$ respectively. 
We also tried to model both distributions in one unified Transformer, but no better performance was observed.
Both $\mathcal{M}_{s}^k$ and $\mathcal{M}_{r}^q$ are trained in a supervised manner upon the offline dataset.  
We use mean-squared error for both the next state prediction and reward prediction.

\textbf{Cyclic annealing schedule.} 
To further enhance the stability, an ensemble of environment models is involved to generate the simulation. 
Most existing ensemble-based methods independently train individual models by setting different initial random seeds, which can be computationally resource-consuming\cite{huang2017snapshot, lu2021revisiting, kidambi2020morel}.
In contrast, we adopt a training efficiency approach to obtain an ensemble of environment models from a single training process.
Specifically, we divide the training process into multiple cycles.
Each cycle begins with a relatively large learning rate, which is subsequently gradually reduced to a smaller value through annealing.
Learning rate decay enables the model to rapidly converge to a local minimum, while a larger learning rate allows the model to escape from a critical point.
Besides, to stabilize training and improve convergence, we initially employed the warmup technique before the first cycle.
The cyclic annealing learning rate schedule can be defined as:
\begin{equation}\label{equ:lr}
\alpha(t) = 
\begin{cases}
\frac{(t + 1)\alpha_0}{T_{\text{wp}}} & \text{if} \quad t < T_{\text{wp}} \\
\frac{\alpha_0}{2} \left( 1 + \cos\left( \pi \cdot \frac{(t - T_{\text{wp}}) \mod T_{\text{cyc}}}{T_{\text{cyc}}} \right) \right) & \text{if} \quad t \geq T_{\text{wp}}
\end{cases}
\end{equation}
where $T_{\text{wp}}$ and $T_{\text{cyc}}$ are hyperparameters representing the number of warmup steps and steps in each cycle, respectively.
At the end of each training cycle, we save the model weights as an individual model of the ensemble.
Once a predefined number of environment models have been obtained, the state and reward prediction is formulated as the mean of the ensemble: 
$\hat{s}_{t+1}=\overline{\hat{s}_{t+1}^k}$, and $\hat{r}_{t}=\overline{\hat{r}_{t}^q}$.

\subsection{Trajectory generation}\label{subsec:trajectorygen}

We now discuss how to generate trajectories with World Transformers. 
The key idea in trajectory generation is to expand the observed state-action tuples, thereby improving the performance of offline RL. 
Motivated by this, we introduce uniform noise to the action $a_{t}$ in the original trajectory, forming a perturbed action $a^*_{t}$. 
In particular, we add a random number of range $(-\epsilon,\epsilon)$ to each dimension of $a_{t}$. 
Then we predict the corresponding next state and reward with World Transformers. 
The alteration of action results in changes to the next state, therefore a new tuple $\{a^*_{t},\hat{r}_{t},\hat{s}_{t+1}\}$ is generated. 
We repeat this process of introducing noise to the original action and predicting the next state and reward from timestep $t=t_{s}$ to $t=t_{s}+h-1$, where $h$  defines the length of simulation. 

Generating simulation for all timesteps in the offline dataset is not appropriate since (1) this may introduce extra noise and (2) not every tuple $\{s_{t},a_{t},r_{t}\}$ in a trajectory is worthy of augmenting. We prefer to generate trajectories with high total rewards, which are good demonstrations for policy improvement. In this paper, three trajectory generation strategies are proposed: \\ \textbf{Random :} For each trajectory $\tau=\{s_{t},a_{t},r_{t}\}^{|\tau|}_{t=0}$, we randomly select the start timestep $t_{s}\in[0, |\tau|-h+1]$, and generate trajectory $\tau^*$ based on the trajectory segment $\{s_{t},a_{t},r_{t}\}^{t_{s}+h-1}_{t=t_{s}}$.\\
\textbf{Top-$N$ :} We first split all original trajectories into segments of length $h$. Then we sort these segments by their cumulative rewards in descending order and generate trajectories based on the Top-$N$ segments, where $N=|\mathcal{D}_{aug}|/h$. \\
\textbf{Softmax :} We also split the trajectory segments as the Top-$N$ strategy, but we choose the segments according to their probabilities calculated by a softmax function over segment  cumulative reward. \\


\setlength{\tabcolsep}{1pt} 
\renewcommand\arraystretch{1.1}
\begin{table*}[t]
    
    \caption{
    Performance comparison when integrating OTTO with existing offline RL methods on Gym MuJoCo-v2. The normalized scores are computed over 5 random seeds. Med-Exp represents \textit{medium-expert} and Med-Rep represents \textit{medium-replay}. 
    Boldface denotes the improved performance after integration.}
    \centering
    
    \resizebox{\linewidth}{!}
    {
        \begin{tabular}{l|l|cc|cc|cc|cc|c|c|c}
            \hline
            \multirow{2}*{\textbf{}} & \multirow{2}*{\textbf{Environment}} & \multicolumn{2}{c|}{\textbf{IQL}} & \multicolumn{2}{c|}{\textbf{TD3+BC}} & \multicolumn{2}{c|}{\textbf{DT}} & \multicolumn{2}{c|}{\textbf{CQL}} & \multirow{2}*{\textbf{COMBO}} & \multirow{2}*{\textbf{BooT}} & \multirow{2}*{\textbf{MOPO}} \\
            \cline{3-10}
            &     & Original & \textbf{OTTO}  & Original  & \textbf{OTTO}  & Original & \textbf{OTTO} & Original & \textbf{OTTO} &  &  &  \\
            \hline
           \multirow{3}{*}{ {\rotatebox{90}{Med-Exp}}} 
            & HalfCheetah     & 92.1{$\pm$1.2} & \textbf{93.8}{$\pm$0.6} 
                              & 93{$\pm$1.2}   & \textbf{95.6}{$\pm$0.3}
                              & 87.7{$\pm$1.4} & \textbf{92.8}{$\pm$1.9}
                              & 89.9{$\pm$2.8} & \textbf{91.4}{$\pm$0.5}
                              & 90 & 94 & 63.3\\
            & Hopper          & 102{$\pm$5.6}   & \textbf{113.4}{$\pm$0.3} 
                              & 101.6{$\pm$2.9} & \textbf{110.6}{$\pm$1.4} 
                              & 105.3{$\pm$6}   & \textbf{110.3}{$\pm$0.8}
                              & 99.1{$\pm$4.3}  & \textbf{107.1}{$\pm$1.3}  
                              & 111.1 & 102.3 & 23.7\\
            & Walker2d        & 111.4{$\pm$0.5} & \textbf{112.9}{$\pm$0.1} 
                              & 109.4{$\pm$0.1} & \textbf{110.9}{$\pm$0.1} 
                              & 108.1{$\pm$0.7} & \textbf{108.9}{$\pm$0.3}           
                              & 109.2{$\pm$0.1} & \textbf{110.5}{$\pm$0.1}
                              & 103.3 & 110.4 & 44.6\\
            \hline
           \multirow{3}{*}{ {\rotatebox{90}{Medium}}} 
           & HalfCheetah      & 48.5{$\pm$0.3} & \textbf{49.3}{$\pm$0.1} 
                              & 48.5{$\pm$0.1} & \textbf{49.9}{$\pm$0.2}
                              & 41.4{$\pm$0.1} & \textbf{44.6}{$\pm$0.2}                    
                              & 46.8{$\pm$0.1} & \textbf{48.1}{$\pm$0.1}
                              & 54.2 & 50.6 & 42.3\\
            & Hopper          & 61.2{$\pm$4.7} & \textbf{78.6}{$\pm$3.5} 
                              & 60.6{$\pm$1} & \textbf{74.5}{$\pm$3.7} 
                              & 64.9{$\pm$4.6} & \textbf{74.2}{$\pm$1.3}
                              & 58.6{$\pm$0.8} & \textbf{62.2}{$\pm$0.7}
                              & 97.2 & 70.2 & 28\\
            & Walker2d        & 80.4{$\pm$0.1} & \textbf{83.5}{$\pm$1.5} 
                              & 81.8{$\pm$0.6} & \textbf{83.7}{$\pm$0.1} 
                              & 73.9{$\pm$2.7} & \textbf{77.1}{$\pm$0.1}              
                              & 81.3{$\pm$0.4} & \textbf{82.7}{$\pm$0.3}
                              & 81.9 & 82.9 & 17.8\\
            \hline
            \multirow{3}{*}{ {\rotatebox{90}{Med-Rep}}}
            & HalfCheetah     & 43.8{$\pm$0.2} & \textbf{44.8}{$\pm$0.1} 
                              & 44.9{$\pm$0.2} & \textbf{45.8}{$\pm$0.2} 
                              & 35.5{$\pm$0.4} & \textbf{38.8}{$\pm$0.5}
                              & 45.4{$\pm$0.3} & \textbf{46.3}{$\pm$0.2}
                              & 55.1 & 46.5 & 53.1\\
            & Hopper          & 94.3{$\pm$4.6} & \textbf{102.4}{$\pm$1} 
                              & 72.8{$\pm$4.9} & \textbf{80.8}{$\pm$13.2} 
                              & 71.2{$\pm$3.6} & \textbf{89.5}{$\pm$5}             
                              & 88.4{$\pm$7.6} & \textbf{92.9}{$\pm$15.4}
                              & 89.5 & 92.9 & 67.5\\
            & Walker2d        & 86.2{$\pm$3.6} & \textbf{86.9}{$\pm$2} 
                              & 87.3{$\pm$1.9} & \textbf{90.7}{$\pm$0.3} 
                              & 70.2{$\pm$1.2} & \textbf{75.8}{$\pm$2.6}               
                              & 78.7{$\pm$1.1} & \textbf{91.1}{$\pm$0.4}
                              & 56.0 & 87.6 & 39\\
            \hline
            \multicolumn{2}{l|}{MuJoCo Average} & 80   & \textbf{85.1}
                                                & 77.8 & \textbf{82.5}
                                                & 73.1 & \textbf{79.1}
                                                & 77.5 & \textbf{81.4}
                                                & 82 & 81.9 & 42.1 \\
        
            \hline
        \end{tabular}
    }
    \label{tab:main_model_free_base}
\end{table*}

\vspace{-10pt}
\subsection{World evaluator}\label{subsec:worldevaluator}
Although the high-reward trajectories could be generated by the manually designed strategies in section \ref{subsec:trajectorygen}, the generated trajectories might be overestimated, especially when World Transformers have high uncertainty regarding their predictions. 
This overestimation might also mislead the policy learning, resulting in sub-optimal performance. 
To address this issue, the World Evaluator is introduced to evaluate the confidence of generated trajectories and give a conservative punishment for trajectories with high uncertainty. 

The uncertainty evaluation is based on variances of both state predictions $\hat{s}_{t+1}^k$ and reward predictions $\hat{r}_{t}^q$. 
Specifically, for each tuple $(a^*_{t},\hat{r}_{t},\hat{s}_{t+1})$ in generated trajectories, we calculate the standard deviation, i.e., ${\sigma}^s_{t+1}$ and ${\sigma}^r_t$, over $\hat{s}_{t+1}^k$ and $\hat{r}_{t}^q$. 
Higher deviation denotes higher uncertainty about trajectory generation, and thus the predicted reward should be downgraded.
Then we correct $
\hat{r}_{t}$ as:
\begin{equation}\label{equ:correct}
r^*_{t}=(1-\frac{\mathrm{e}^{(\sigma^s_{t+1}/\omega)}}{\sum_{t=t_{s}}^{t_{s}+h-1}\mathrm{e}^{(\sigma^s_{t+1}/\omega)}})(\hat{r}_{t}-{\sigma}^r_t),
\end{equation}
where $\omega$ is a hyperparameter representing the temperature. $\hat{r}_{t}-{\sigma}^r_t$ ensures a lower bound of the predicted reward and $1-\frac{\mathrm{e}^{(\sigma^s_{t+1}/\omega)}}{\sum_{t=t_{s}}^{t_{s}+h-1}\mathrm{e}^{(\sigma^s_{t+1}/\omega)}}$ evaluates the confidence level of the predicted next states. If World Transformers are not confident about the state transition and reward, a conservative punishment will be introduced to downgrade the predicted reward. 
Finally, the tuple $(a_t^*,r_t^*, \hat{s}_{t+1})$ is augmented to enlarge the offline data. 
The whole process for trajectory generation and correction is visualized in Figure \ref{fig:framework}. After data augmentation, an offline RL algorithm is trained upon the augmented dataset to gain better policy performance.


\section{Experiment}\label{sec:exp}
In this section, we conduct experiments to answer the following research questions: (1) Can OTTO effectively improve the performance of existing offline model-free RL methods? 
(2) How does OTTO perform compared to state-of-the-art data augmentation methods for offline RL? 
(3) How does the 
design of OTTO affect the policy performance?

\setlength{\tabcolsep}{3pt} 
\begin{table}[t]
    \caption{The average score of AntMaze. 
    The normalized scores are computed over 5 random seeds. 
    Boldface denotes the highest score.
    }
    \centering
    \begin{tabular}{l|cc|cc}
        \hline
        \multirow{2}*{\textbf{Dataset}}  &  \multicolumn{2}{c|}{\textbf{IQL}}  &  \multicolumn{2}{c}{\textbf{TD3+BC}} \\
        \cline{2-5} 
         &    Original  &  OTTO  &  Original  &  OTTO \\
        \hline
        umaze          & \textbf{73.6}{$\pm$2.8} & 73{$\pm$2.1} 
                       & 93.1{$\pm$1.4}   & \textbf{95.7}{$\pm$2.8}\\
        umaze-diverse  & 58{$\pm$2}     & \textbf{60.4}{$\pm$2.3} 
                       & 40.5{$\pm$3.9} & \textbf{53.3}{$\pm$5.4}\\
        medium-play    & 67.3{$\pm$1.9} & \textbf{72.4}{$\pm$3.2} 
                       & 0.4{$\pm$0.2} & \textbf{2.1}{$\pm$0.2}\\
        medium-diverse & \textbf{68.4}{$\pm$1.8} & 67.8{$\pm$2.5} 
                       & 0.3{$\pm$0.1} & \textbf{0.5}{$\pm$0.1}\\
        large-play     & 41{$\pm$7.4}   & \textbf{43.2}{$\pm$3.3} 
                       & 0{$\pm$0} & 0{$\pm$0}\\
        large-diverse  & 31.2{$\pm$2.6} & \textbf{36.8}{$\pm$2.1} 
                       & 0{$\pm$0} & 0{$\pm$0}\\
        \hline
        Antmaze Average  & 56.6 & \textbf{58.9} & 22.3 & \textbf{25.3} \\
        \hline
    \end{tabular}
\label{tab:ant_maze}
\vspace{-8pt}
\end{table}
\setlength{\tabcolsep}{0pt} 
\setlength{\tabcolsep}{2pt} 
\begin{table*}[t]
    \caption{
    Comparison with diffusion-based augmentation baselines. The normalized scores are computed over 5 random seeds. Boldface denotes the highest score.
    }
    \centering
    {
        \begin{tabular}{ll|ccc|ccc|cc}
            \hline
            \multirow{2}*{\textbf{Dataset}} & \multirow{2}*{\textbf{Environment}}&\multicolumn{3}{c|}{\textbf{IQL}} & \multicolumn{3}{c|}{\textbf{TD3+BC}} & \multicolumn{2}{c}{\textbf{DT}}\\
            \cline{3-10}
            &     & SER  & DStitch   & \textbf{OTTO}  & SER  & DStitch   & \textbf{OTTO} & DStitch & \textbf{OTTO} \\
            \hline
            Med-Expert & HalfCheetah     & 88.9 & \textbf{94.4} & 93.8{$\pm$0.6} 
                                         & 86.5 & \textbf{96} & 95.6{$\pm$0.3} 
                                         & 92.6 & \textbf{92.8}{$\pm$1.9}\\
            Med-Expert & Hopper          & 110.4 & 110.9 & \textbf{113.4}{$\pm$0.3} 
                                         & 104 & 107.1 & \textbf{110.6}{$\pm$1.4} 
                                         & 109.4 & \textbf{110.3}{$\pm$0.8}\\
            Med-Expert & Walker2d        & 111.7 & 111.6 & \textbf{112.9}{$\pm$0.1} 
                                         & 110.5 & 110.2 & \textbf{110.9}{$\pm$0.1} 
                                         & 108.6 & \textbf{108.9}{$\pm$0.3}\\
            \hline
            Medium & HalfCheetah      & 49.3 & \textbf{49.4} & 49.3{$\pm$0.1} 
                                      & 48.4 & \textbf{50.4} & 49.9{$\pm$0.2} 
                                      & 44.2 & \textbf{44.6}{$\pm$0.2}\\
            Medium & Hopper           & 66.6 & 71 & \textbf{78.6}{$\pm$3.5} 
                                      & 56.4 & 60.3 & \textbf{74.5}{$\pm$3.7} 
                                      & 60.5 & \textbf{74.2}{$\pm$1.3}\\
            Medium & Walker2d         & \textbf{85.9} & 83.2 & 83.5{$\pm$1.5} 
                                      & \textbf{84.9} & 83.4 & 83.7{$\pm$0.1} 
                                      & 72 & \textbf{77.1}{$\pm$0.1}\\
            \hline
            Med-Replay & HalfCheetah     & 46.6 & 44.7 & \textbf{44.8}{$\pm$0.1} 
                                         & 45.2 & 44.7 & \textbf{45.8}{$\pm$0.2} 
                                         & \textbf{41} & 38.8{$\pm$0.5}\\
            Med-Replay & Hopper          & 102.4 & 102.1 & \textbf{102.4}{$\pm$1} 
                                         & 56.8 & 79.6 & \textbf{80.8}{$\pm$13.2} 
                                         & \textbf{96.1} & 89.5{$\pm$5}\\
            Med-Replay & Walker2d        & 85.7 & 86 & \textbf{86.9}{$\pm$2} 
                                         & 89.1 & 89.7 & \textbf{90.7}{$\pm$0.3} 
                                         & 60.2 & \textbf{75.8}{$\pm$2.6}\\
            \hline
            \multicolumn{2}{l|}{MuJoCo Average} & 83.1 & 83.8 & \textbf{85.1} 
                                                & 75.8 & 80.2 & \textbf{82.5} 
                                             & 76.1 & \textbf{79.1}\\

            \hline
        \end{tabular}
    }
    \label{tab:result_iql_td3_dt}
\end{table*}
\setlength{\tabcolsep}{0pt} 
\textbf{Baselines.} To answer question (1), we integrate OTTO with various representative model-free offline RL approaches and compare the policy performance with or without OTTO. These model-free offline RL approaches include DT \cite{chen2021decision}, TD3+BC \cite{fujimoto2021minimalist}, CQL \cite{kumar2020conservative} and IQL \cite{kostrikov2021offlineimplicit}. DT casts offline RL into a sequence modeling problem. TD3+BC is an imitation learning method, which jointly learns the policy and value function. CQL and IQL are both based on TD learning, where CQL is a multi-step dynamic programming method and IQL is a one-step method. Besides, we also compare OTTO with three representative model-based offline RL approaches (COMBO \cite{yu2021combo}, BooT \cite{wang2022bootstrapped} and MOPO \cite{yu2020mopo}).
To answer question (2), we compare OTTO with two recently proposed diffusion-based augmentation methods, i.e., SER \cite{lu2024synthetic} and DStitch \cite{li2024diffstitch}.

\textbf{Datasets.} We evaluate our methods on a wide range of datasets in Gym MuJoCo and AntMaze tasks from D4RL benchmark\cite{fu2020d4rl}. Specifically, we use three different environments in Gym MuJoCo domains (\textit{HalfCheetah}, \textit{Hopper}, \textit{Walker2d}), each with three datasets (\textit{Medium}, \textit{Medium-Replay}, \textit{Medium-Expert}). We also evaluate the performance in AntMaze tasks, which aim to control a robot and navigate to reach a goal. There are three different size of maze (\textit{umaze}, \textit{medium}, \textit{large}) and different dataset types (\textit{fixed}, \textit{play}, \textit{diverse}). In AntMaze tasks, the agent receives a sparse reward only if the goal is reached. Besides, it has a larger state space (27 dimensions) and action space (8 dimensions), making it more challenging compared to Gym MuJoCo tasks. 

\textbf{Implementation details.} For World Transformers, the number of State Transformers $K$ the number of Reward Transformers $Q$ are both set to 4. 
The interaction steps of input for the State Transformers $L_s$ and Reward Transformers $L_r$ are both set to 20. 
For trajectory generation, the horizon of generation is set to 50 for both MuJoCo and AntMaze tasks. The temperature $\omega$ of World Evaluator is set to 0.7. 
We set the augmentation ratio $\delta$ ranging from 5\% to 20\% depending on the dataset. 
The number of warmup steps $T_{\text{wp}}$ and the number of steps in each cycle $T_{\text{cyc}}$ are set to $10^5$ and $5 \times 10^5$, respectively.
We need four cycles in total to get four different models.
The full experimental details are provided in Appendix \ref{sec:TrainingDetails}.

\subsection{Performance with OTTO augmentation}\label{subsec:performance_otto}
To answer the RQ1, we evaluate the performance of model-free offline RL methods with the integration of OTTO on both standard MuJoCo tasks and challenging AntMaze tasks.

\textbf{Evaluation on MuJoCo tasks.} Table \ref{tab:main_model_free_base} shows the 
results of four representative model-free offline RL methods (IQL, TD3+BC, DT, CQL) on MuJoCo environment. We also include three representative model-based offline RL approaches (COMBO, BooT, MOPO) for a more comprehensive comparison. 
From Table \ref{tab:main_model_free_base} we can observe that: (1) OTTO significantly outperforms the original scores in all settings, achieving consistent improvement in MuJoCo average score among four representative model-free methods. 
This suggests that OTTO is robust to different datasets and can generalize to multiple model-free algorithms. 
(2) OTTO achieves significant improvement in the datasets where the original score is relatively low such as \textit{Hopper-Medium} datasets. 
This observation suggests that OTTO can potentially generate good demonstrations in datasets that do not include expert data. 
(3) The performance of COMBO and BooT was initially superior to that of model-free RL algorithms. 
After the augmentation of OTTO, two of four model-free RL algorithms (IQL and TD3+BC) outperform COMBO and BooT and the remaining two methods achieve comparable results. 
All four model-free RL algorithms achieve better performance than MOPO after OTTO augmentation. 
This indicates that the long-horizon high-confidence trajectories generated by OTTO provide more significant generalization improvement than the short-horizon rollouts generated by existing model-based methods. 

\textbf{Evaluation on AntMaze tasks.}
We also conduct experiments on the challenging AntMaze tasks and present the results in Table \ref{tab:ant_maze}. 
We can observe that OTTO improves the policy performance in most cases, and obtains the best average score for both IQL and TD3+BC. 
AntMaze is a sparse-reward environment that requires “stitching” parts of sub-optimal trajectories \cite{kostrikov2021offlineimplicit}.  
This indicates that OTTO can generalize between different states and generate trajectories from different start points to the goal point. 
This observation shows the effectiveness of OTTO in complex environments.

To summarize, OTTO effectively improves the performance of various existing model-free RL algorithms across different environments including complex environments like AntMaze.

\setlength{\tabcolsep}{4pt} 
\begin{table*}[t]

    \caption{
    Results of IQL and DT with three trajectory generation strategies of OTTO. The normalized scores are computed over 5 random seeds. Boldface denotes the highest score. 
    }
    \centering
    {
        \begin{tabular}{l l | c c c c| c c c c}
            \hline
            \multirow{2}*{\textbf{Dataset}} & \multirow{2}*{\textbf{Environment}}&\multicolumn{4}{c|}{\textbf{IQL}} & \multicolumn{4}{c}{\textbf{DT}} \\
            \cline{3-10}
            &     & Original      & Random  & Top-$N$   & Softmax  & Original      & Random  & Top-$N$   & Softmax  \\
            \hline
           Med-Expert & HalfCheetah     & 92.1{$\pm$1.2}  & 93.3{$\pm$0.3}  & \textbf{93.8}{$\pm$0.6}  & 91.1{$\pm$0.3} 
                                        & 87.7{$\pm$1.4}  & 91.4{$\pm$0.9}  & \textbf{92.8}{$\pm$1.9}  & 92.5{$\pm$1.8}  \\
           Med-Expert & Hopper          & 102{$\pm$5.6}   & 111.3{$\pm$0.2} & 112.6{$\pm$0.8} & \textbf{113.4}{$\pm$0.3} 
                                        & 105.3{$\pm$6}   & 109.7{$\pm$0.4} & \textbf{110.3}{$\pm$0.8} & 110.2{$\pm$1.4}   \\
           Med-Expert & Walker2d        & 111.4{$\pm$0.5} & 112.5{$\pm$0.1} & 112.8{$\pm$0.3} & \textbf{112.9}{$\pm$0.1} 
                                        & 108.1{$\pm$0.7} & 108.6{$\pm$0.1} & \textbf{108.9}{$\pm$0.3} & 108.5{$\pm$0.2}  \\
            \hline
           Medium & HalfCheetah         & 48.5{$\pm$0.3}  & 48.6{$\pm$0.1}  & 48.5{$\pm$0.3}  & \textbf{49.3}{$\pm$0.1} 
                                        & 41.4{$\pm$0.1}  & 43.5{$\pm$0.1}  & 44{$\pm$1.1}  & \textbf{44.6}{$\pm$0.2}  \\
            Medium & Hopper             & 61.2{$\pm$4.3}  & 72.3{$\pm$2.7}  & 69.7{$\pm$4.6}  & \textbf{78.6}{$\pm$3.5} 
                                        & 64.9{$\pm$4.6}  & 71{$\pm$1.5}  & 69.4{$\pm$2.4}  & \textbf{74.2}{$\pm$1.3}  \\
            Medium & Walker2d           & 80.4{$\pm$0.1}  & 82.3{$\pm$0.4}  & 82.1{$\pm$1.8}  & \textbf{83.5}{$\pm$1.5} 
                                        & 73.9{$\pm$2.7}  & \textbf{77.1}{$\pm$0.1}  & 77{$\pm$1.2}  & 76.9{$\pm$1.3}  \\
            \hline
            Med-Replay& HalfCheetah     & 43.8{$\pm$0.2}  & 44.7{$\pm$0.1}    & 44.3{$\pm$0.1}  & \textbf{44.8}{$\pm$0.1} 
                                        & 35.5{$\pm$0.4}  & 37{$\pm$0.6}    & \textbf{38.8}{$\pm$0.5}  & 38.6{$\pm$0.7}  \\
             Med-Replay& Hopper         & 94.3{$\pm$4.6}  & 100.4{$\pm$1.2}   & 99.8{$\pm$2.3}  & \textbf{102.4}{$\pm$1} 
                                        & 71.2{$\pm$3.6}  & 87.7{$\pm$3.4}  & 89.4{$\pm$4.6}  & \textbf{89.5}{$\pm$5}  \\
             Med-Replay& Walker2d       & 86.2{$\pm$3.6}  & 86.4{$\pm$1.5}  & 86.7{$\pm$2.8}  & \textbf{86.9}{$\pm$2} 
                                        & 70.2{$\pm$1.2}  & 72.3{$\pm$1.5}  & 71{$\pm$1.6}    & \textbf{75.8}{$\pm$2.6}  \\
            \hline
            \multicolumn{2}{l |}{MuJoCo-v2 Average} & 80 & 83.5 & 83.4 & \textbf{84.8} 
                                                    & 73.1 & 77.6 & 78 & \textbf{79} \\

            \hline
        \end{tabular}
    }
    \label{tab:three_strategies_with_iql_dt}
\end{table*}
\setlength{\tabcolsep}{0pt}

\subsection{Comparison with augmentation methods}\label{comparison_da}
To answer RQ2, we compared OTTO with the state-of-the-art diffusion-based data augmentation algorithms\cite{li2024diffstitch,lu2024synthetic}.

Table \ref{tab:result_iql_td3_dt} shows the comparison between OTTO and state-of-the-art diffusion-based data augmentation baselines (SER, DStitch) on Gym MuJoCo tasks. 
The result of SER for DT is omitted since SER cannot be utilized for sequence modeling-based DT \cite{li2024diffstitch}.
From Table \ref{tab:result_iql_td3_dt} we can observe that OTTO achieves better scores than SER and DStitch in 18 out of 27 settings, and achieves comparable results in the remaining settings.
The reason is that although SER and DStitch utilize the powerful diffusion model to generate trajectory augmentation, they do not conduct the evaluation and correction for generated trajectories. 
On the contrary, OTTO introduces an uncertainty-based World Evaluator to evaluate the generated data and uses a conservative punishment to correct low-confidence trajectories, leading to higher-confidence augmentation. 
This observation verifies the effectiveness of the World Evaluator.

\subsection{Method investigation}
\subsubsection{Effect of trajectory generation strategies}\label{subsec:effect_strategies}


In this section, we evaluate the three trajectory generation strategies on Gym MuJoCo tasks. Table \ref{tab:three_strategies_with_iql_dt} summarizes the experimental results for IQL and DT. 
Experiment results in Table \ref{tab:three_strategies_with_iql_dt} show that all of the three strategies can outperform the original performance, which demonstrates that all of these strategies can benefit the existing offline RL algorithms. 

In addition, for each strategy, we have the following key observations: 
(1) The Random strategy exhibits lower standard deviation, indicating that augmenting trajectories through random selection preserves the distribution of the augmented trajectories in alignment with the original dataset.
As a result, this strategy produces more stable outcomes and demonstrates reduced standard deviation.
(2) The Top-$N$ strategy achieves better performance when there are expert data in the dataset (\textit{Medium-Expert}), demonstrating its ability to effectively select high-reward trajectories for augmentation.
(3) The Softmax strategy achieves the best score in most cases, demonstrating that it not only generates high-reward trajectories but also enhances robustness.
Based on the aforementioned observations, we provide guidance on how to select an appropriate strategy according to the given environment.
To ensure a lower bound on performance, the Random strategy can be selected to obtain relatively stable outputs, as it exhibits the smallest standard deviation. 
If expert data is present in the environment, the Top-$N$ strategy should be chosen to better leverage the expert knowledge. 
In other cases, or when facing an unknown environment, the Softmax strategy is recommended.


\subsubsection{Ablation study}


To investigate the impact of World Transformers and World Evaluator, we use IQL to conduct an ablation study on Gym-MuJoCo tasks. The results are presented in Table \ref{tab:ablation}. Specifically, we have the variants:

\textbf{Original}: train the offline RL algorithms on the original dataset.

\textbf{Single}: only use a single $\mathcal{M}_{r}$ and a a single $\mathcal{M}_{s}$ to predict the reward and next states, respectively.

\textbf{No-Correct}: use an ensemble of Reward Transformers and State Transformers to predict the reward and next states, but remove the correction of World Evaluator.

\begin{table}[t]
    \caption{Ablation study. The normalized scores are computed over 5 random seeds.  Boldface denotes the highest score.
    }
    \centering
    {
    \begin{tabular}{l|cccc}
        \hline
        \textbf{Dataset}    &\textbf{Original} &\textbf{Single} &\textbf{No-Correct} &\textbf{OTTO}  \\
        \hline
        Med-Expert-Halfcheetah &   92.1$\pm$1.2  & 82.5$\pm$2.8  & 84.2$\pm$2.2  & \textbf{93.8}$\pm$0.6\\
        Med-Expert-Hopper      &   102$\pm$5.6   & 105.9$\pm$1.6 & 111.9$\pm$0.2 & \textbf{113.4}$\pm$0.3\\
        Med-Expert-Walker2d    &   111.4$\pm$0.5 & 111.5$\pm$0.1 & 111.5$\pm$0.5 & \textbf{112.9}$\pm$0.1\\
        \hline
        Medium-Halfcheetah        &   48.5$\pm$0.3  & 47.5$\pm$0.1  & 48.5$\pm$0.4  & \textbf{49.2}$\pm$0.1\\
        Medium-Hopper             &   61.2$\pm$4.3  & 66.6$\pm$14.2 & 77.9$\pm$8.5   & \textbf{78.6}$\pm$3.5\\
        Medium-Walker2d           &   80.4$\pm$0.1  & 78.3$\pm$0.5  & 82.7$\pm$2.3  & \textbf{83.5}$\pm$1.5\\
        \hline
        Med-Replay-Halfcheetah &   43.8$\pm$0.2  & 44.1$\pm$0.1  & 43.3$\pm$0.5  & \textbf{44.8}$\pm$0.1\\
        Med-Replay-Hopper      &   94.3$\pm$4.6  & 90.9$\pm$15.5 & 100.6$\pm$0.6 & \textbf{102.4}$\pm$1\\
        Med-Replay-Walker2d    &   86.2$\pm$3.6  & 83.7$\pm$14.7 & 83.4.2$\pm$0.8  & \textbf{86.9}$\pm$2\\
        \hline
        {MuJoCo Average}      &80  & 79.5 & 82.7  & \textbf{85.1}\\
        \hline
    \end{tabular}
    }
    \vspace{-15pt}
\label{tab:ablation}
\end{table}

From Table \ref{tab:ablation} we observe that: (1)
Single and Original show comparable results and the score standard deviation of Single is higher, especially in \textit{Med-Replay-Hopper} ($\pm$15.5) and \textit{Med-Replay-Walker2d} ($\pm$14.7), which suggests that the trajectories generated by using a single State Transformer and Reward Transformer could be unreliable. 
(2) No-Correct outperforms Original and Single in most cases, which demonstrates the necessity of utilizing an ensemble of Reward Transformers and State Transformers to model the environment dynamics.
(3) OTTO achieves the best performance  in most cases, indicating that the correction of the World Evaluator effectively improves policy learning. 

Besides, we also evaluate the quality of trajectories by calculating the accuracy of the predicted rewards of Single and OTTO, the results are presented in Appendix \ref{additional_exp}. 
We can observe that OTTO makes more accurate prediction than Single, indicating that the correction of the World Evaluator improves the quality of generated trajectories. 
To summarize, the ensemble of World Transformers together with the World Evaluator are both essential to improve the offline RL performance.

\subsubsection{Hyperparameter study}\label{subsec:hyperparameter_study}
\begin{figure}[t]
\begin{center}
\centerline{\includegraphics[width=0.9\columnwidth]{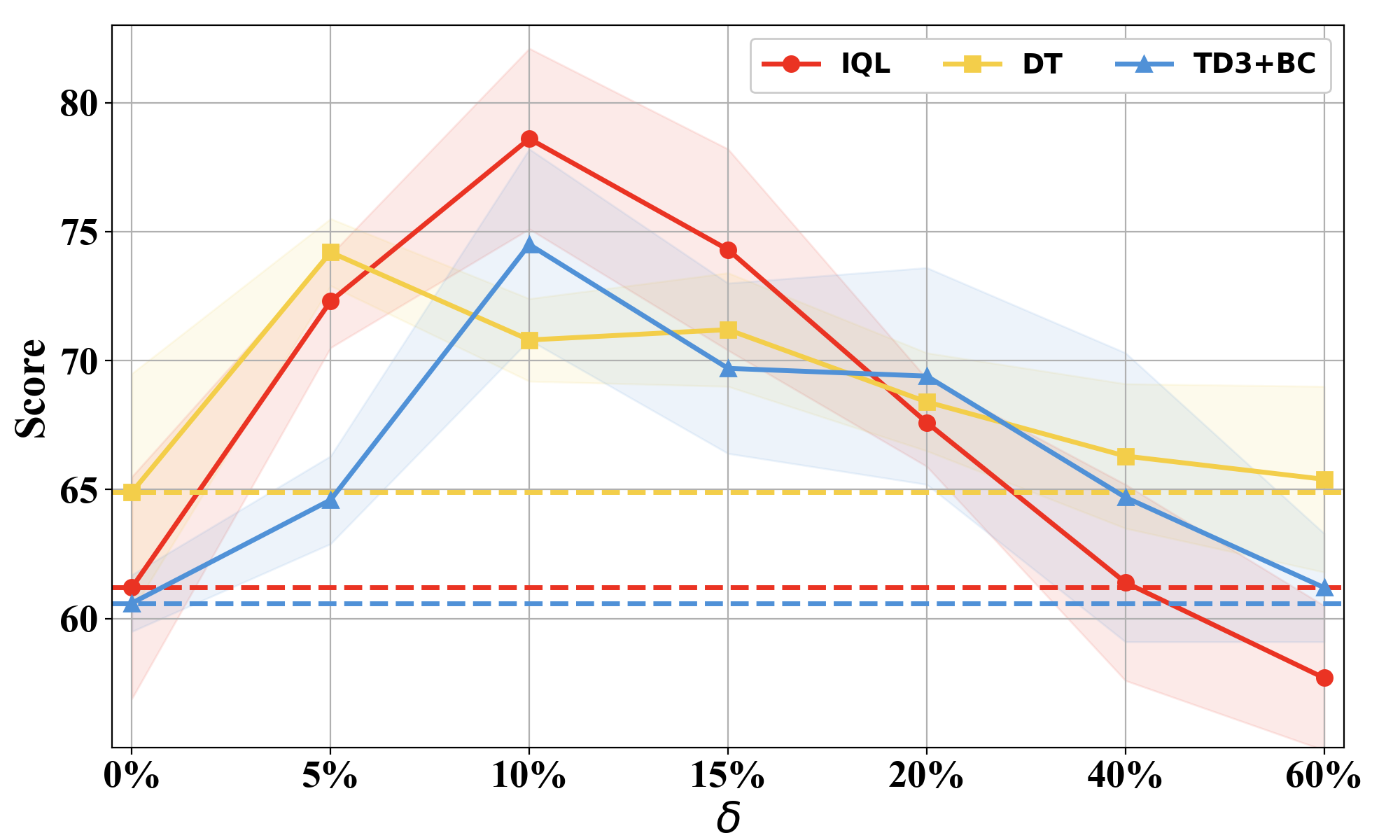}}
\caption{Impact of $\delta$. Dashed lines are original scores.}
\label{fig:delta}
\end{center}
\end{figure}

In this section, we perform a hyperparameter study on augmentation ratio $\delta$, noise range $\epsilon$, and the number of models $K$ and $Q$.

\textbf{Impact of augmentation ratio $\delta$}. To further investigate the impact of augmentation ratio $\delta$, we conduct experiments by evaluating IQL on \textit{Hopper-Medium} dataset with different augmentation ratios $\delta$. Specifically, we set the augmentation ratio from 5\% to 60\% and present the results in Figure \ref{fig:delta}. We can observe that the performance of all three offline RL algorithms first increases and then decreases. When the augmentation ratio increases at the beginning, the performance of OTTO improves because the mixed dataset with more augmented data is able to provide more high-reward trajectories to enlarge the observed states and actions. However, when the augmentation ratio is too large, the performance will drop because too much noise could be introduced to the dataset.

\begin{table}[t]
\setlength{\tabcolsep}{2pt} 
    \caption{Impact of $\epsilon$. The normalized scores are computed over 5 random seeds. Boldface denotes the highest score.
    }
    \centering
    \resizebox{\linewidth}{!}{
    \begin{tabular}{l|cccc}
        \hline
        \textbf{Dataset}  &Original &\textbf{$\bm{\epsilon=0.02}$} &\textbf{$\bm{\epsilon=0.1}$} &\textbf{$\bm{\epsilon=0.5}$}  \\
        \hline
        Med-Expert-Halfcheetah &   92.1{$\pm$1.2}  & \textbf{93.8}{$\pm$0.6}  & 93{$\pm$0.8}  & 92.5{$\pm$1.2}\\
        Med-Expert-Hooper      &   102{$\pm$5.6}   & 112.4{$\pm$1.2} & \textbf{113.4}{$\pm$0.3} & 108.8{$\pm$0.4}\\
        Med-Expert-Walker2d    &   111.4{$\pm$0.5} & \textbf{112.9}{$\pm$0.1} & 112.2{$\pm$0.1}   & 111.7{$\pm$0.3}\\
        \hline
        Med-Halfcheetah        &   48.5{$\pm$0.3}  & \textbf{49.3}{$\pm$0.1}  & 49.1{$\pm$0.1}    & 48.2{$\pm$0.1}\\
        Med-Hooper             &   61.2{$\pm$4.3}  & 76.7{$\pm$2.6}  & \textbf{78.6}{$\pm$3.5}  & 56.3{$\pm$7.2}\\
        Med-Walker2d           &   80.4{$\pm$0.1}  & 82.5{$\pm$3.2}  & \textbf{83.5}{$\pm$2.3}    & 80.7{$\pm$5.2}\\
        \hline
        Med-Replay-Halfcheetah &   43.8{$\pm$0.2}  & \textbf{44.8}{$\pm$0.2}  & \textbf{44.8}{$\pm$0.1}  & 44.5{$\pm$0.9}\\
        Med-Replay-Hooper      &   94.3{$\pm$4.6}  & \textbf{102.4}{$\pm$1} & 100.5{$\pm$2.4}  & 92.3{$\pm$4.6}\\
        Med-Replay-Walker2d    &   86.2{$\pm$3.6}  & 86.8{$\pm$2.4}  & \textbf{86.9}{$\pm$2}  & 73.9{$\pm$5.4}\\
        \hline
        MuJoCo Average      &   80  & 84.6 & \textbf{84.7}  & 78.8\\
        \hline
    \end{tabular}
    }
\label{tab:parameter_epsilon}
\end{table}

\textbf{Impact of noise range $\epsilon$}. We perform the hyperparameter study on $\epsilon$ by evaluating IQL on Gym MuJoCo tasks. Table \ref{tab:parameter_epsilon} shows the results. We can observe that when $\epsilon$ is too large, e.g. $\epsilon=0.5$, the average score is lower than Original. This demonstrates that too large perturbation could downgrade the method's performance. Besides, $\epsilon=0.1$ achieves comparable performance with  $\epsilon=0.02$ and both of them outperform the original score. In simple environments like Hopper, $\epsilon=0.1$ performs better. This indicates that the estimation is more accurate in such simple environments and a larger perturbation range leads to a broader generalization range and a better performance. 

\begin{figure}[t]
\begin{center}
\centerline{\includegraphics[width=0.9\columnwidth]{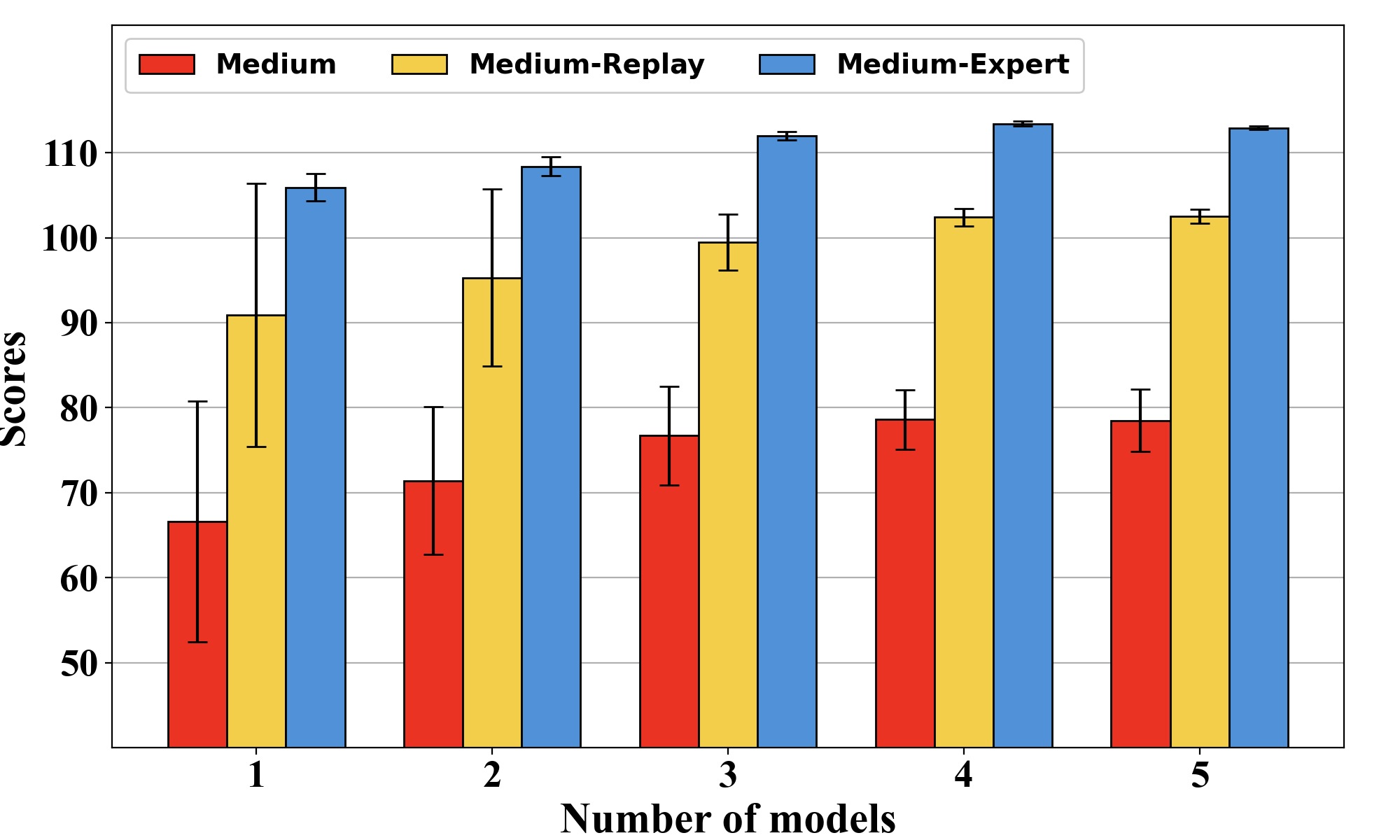}}
\caption{Impact of the number of models $K$ and $Q$.}
\label{fig:numberofmodel}
\end{center}
\end{figure}

\textbf{Impact of the number of models $K$ and $Q$}. We conduct experiments by evaluating IQL on \textit{Hopper} environment with different number of models. 
Specifically, we set the number of State Transformers $K$ equal to the number of Reward Transformers $Q$, and increase this number from 1 to 5.
The results are shown in Figure \ref{fig:numberofmodel}.
We observed that as the number of models increases, the performance also improves, while the standard deviation decreases, indicating that increasing the number of models can lead to more accurate predictions.
In addition, the performance when the number of models is 4 is comparable to that when the number is 5.
Considering that using more models increases training time, we use $K=Q=4$ as a default setting.
\section{Conclusion}
In this work, we have presented offline trajectory optimization for offline reinforcement learning (OTTO). In particular, we have trained  World Transformers to simulate the environment and proposed three generation strategies to generate long-horizon trajectories. Then World Evaluator is introduced to evaluate and correct the generated trajectories.
The RL agent is trained upon both the original data and augmented trajectories.
Extensive experiments on D4RL benchmarks show that OTTO improves the performance of representative model-free offline RL algorithms and outperforms strong model-based baselines. 
For future work, how to improve the World Evaluator and generate a large volume of augmented data to further enhance policy learning is a promising direction. 


\begin{acks}
    This research was (partially) supported by the Natural Science Foundation of China (62202271, 62472261, 62372275, 62272274, T2293773), the National Key R\&D Program of China with grant No. 2024YFC3307300 and No. 2022YFC3303004, the Provincial Key R\&D Program of Shandong Province with grant No. 2024CXGC010108, the Natural Science Foundation of Shandong Province with grant No. ZR2024QF203, the Technology Innovation Guidance Program of Shandong Province with grant No. YDZX2024088.
    All content represents the opinion of the authors, which is not necessarily shared or endorsed by their respective employers and/or sponsors.
\end{acks}

\bibliographystyle{ACM-Reference-Format}
\bibliography{sample-base}

\appendix





\section{Experimental details}\label{sec:TrainingDetails}

\subsection{Hyperpatameter setting}
In this section, we discuss the hyperparameters that we use for OTTO. Table \ref{tb:hyper} lists the hyperparameters we use to train the World Transformers. We use the same hyperparameters for both the State Transformers and Reward Transformers. Both the State Transformer and Reward Transformer are optimized by Adam \cite{kingma2014adam} optimizer.
\begin{table}[h]
\caption{Hyperparameters of World Transformers.}
\label{tb:hyper}
\begin{center}
\begin{small}
\begin{tabular}{ll}
\toprule
\textbf{Hyperparameter}& \textbf{Value}\\
\midrule
Embedding dimension & 128\\
Activation function & ReLU \\
Batch size & 64\\
$L_s,L_r$ (interaction steps of input) & 20\\
$K,Q$ (number of models) & 4\\
Number of layers&10 \\
Number of attention heads & 4 \\
Dropout & 0.1\\
Learning rate & $10^{-4}$\\
Grad norm clip & 0.25\\
Weight decay & $10^{-4}$\\
Number of warmup steps $T_{\text{wp}}$ & $10^5$ \\
Number of steps in each cycle $T_{\text{cyc}}$& 5$\times 10^5$\\
\bottomrule
\end{tabular}
\end{small}
\end{center}

\end{table}

Table \ref{tb:hyper_gen} lists the hyperparameters we use to generate the trajectories.

\setlength{\tabcolsep}{15pt} 
\begin{table}[h]
\caption{Hyperparameters of generating trajectories.}
\label{tb:hyper_gen}
\begin{center}
\begin{small}
\begin{tabular}{ll}
\toprule
\textbf{Hyperparameter}& \textbf{Value (Search Range)}\\
\midrule
augmentation ratio $\delta$ & \{5\%,10\%,15\%,20\%\}\\
range of noise $\epsilon$  & \{0.02, 0.1\} \\
trajectory length $h$ & 50\\
temperature$\omega$ & 0.7\\
\bottomrule
\end{tabular}
\end{small}
\end{center}
\end{table}

\begin{figure*}[t]
\centering
\begin{center}
\centerline{\includegraphics[width=\linewidth]{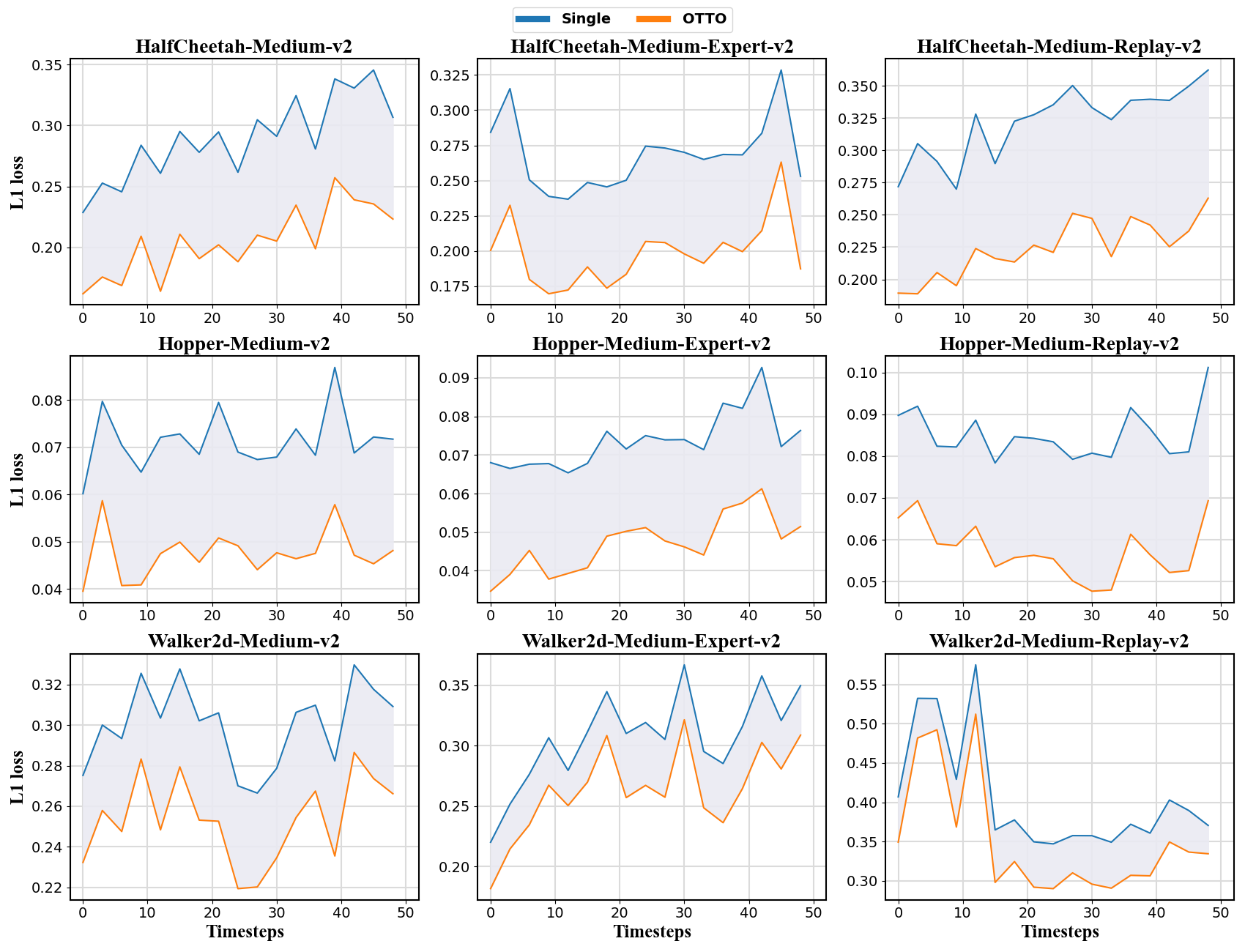}}
\caption{The average L1 loss between the real states and the predicted states of each interaction step in the trajectories generated by Single and OTTO.}
\label{fig:l1loss_state}
\end{center}
\end{figure*}

\subsection{Implementation details}

For a fair comparison, we adopt the same training hyperparameters according to their original papers \cite{chen2021decision,kumar2020conservative,kostrikov2021offlineimplicit,fujimoto2021minimalist} with all the model-free offline RL methods (IQL, DT, CQL, TD3+BC) in all the experiments. 

For IQL, TD3+BC and CQL, we use the CORL (Clean Offline Reinforcement Learning) codebase \cite{tarasov2022corl}, which can be found in \url{https://github.com/tinkoff-ai/CORL} and is released under an Apache-2.0 license. For DT, we use the official codebase, which can be found in \url{https://github.com/kzl/decision-transformer} and is released under an MIT license.

The D4RL dataset we use is released under Creative Commons Attribution 4.0 License (CC BY) License, which can be found in \url{https://github.com/Farama-Foundation/D4RL/tree/master}.

We use NVIDIA GeForce RTX 2080 Ti to train each model. It takes 1.8-2.5 GPUhours to train one single State Transformer or Reward Transformer and 4-8 hours to generate trajectories, which depends on the different datasets.

\section{Additional experiments.} \label{additional_exp}

\begin{figure*}[t]
\centering
\begin{center}
\centerline{\includegraphics[width=\linewidth]{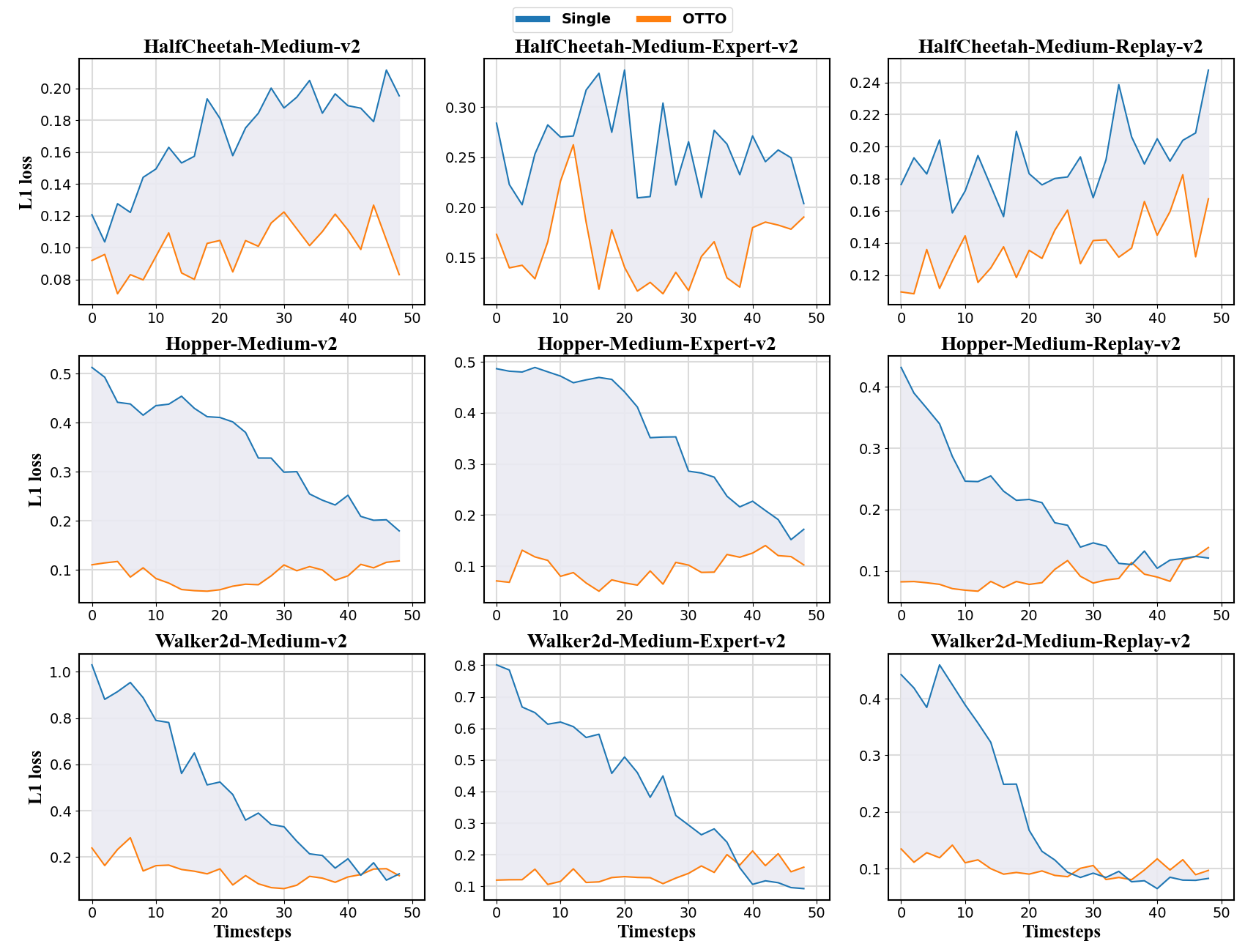}}
\caption{The average L1 loss between the real rewards and the corrected rewards of each interaction step in the trajectories generated by Single and OTTO.}
\label{fig:l1loss_reward}
\end{center}
\end{figure*}

To further investigate the influence of the ensemble of World Transformers, we calculate the average L1 loss of states and reward between the real value and the predicted value of each interaction step in the trajectories generated by Single and OTTO. The results are presented in Figure \ref{fig:l1loss_state} and Figure \ref{fig:l1loss_reward}. We can observe that the predicted accuracy of OTTO is better than Single in most settings, which demonstrates the necessity of utilizing an ensemble of Reward Transformers and State Transformers to model the environment dynamics and utilizing World Evaluator to evaluate and correct the generated trajectories.

\end{document}